\documentclass{article}
\usepackage{arxiv}

\usepackage{graphicx}       
\usepackage{cite}           
\usepackage{xcolor}         
\usepackage{lipsum}         
\usepackage{amsfonts}       
\usepackage{amsmath}        
\usepackage{amssymb}        
\usepackage{alphabeta}      
\usepackage{url}            
\usepackage{nameref}        
\usepackage{cases}          
\usepackage{xfrac}          
\usepackage{algorithm}      
\usepackage{algpseudocodex} 
\usepackage{float}          
\usepackage{anyfontsize}    
\usepackage{titlesec}       
\usepackage{subfig}         
\usepackage{caption}        
\usepackage{abstract}       
\usepackage[shortlabels, inline]{enumitem} 
\usepackage{setspace}       
\usepackage{colortbl}       

\newcommand{\norm}[2]{ \left\Vert {#1} \right\Vert_{#2}}

\newcommand{\paragraphfont}[1]{\textsc{{#1}.}}

\makeatletter
\def\@overkeywordskip{1.5pt} 
\makeatother

\newcommand{\RNN}{RNN}
\newcommand{\LSTM}{LSTM}
\newcommand{\GRU}{GRU}
\newcommand{\SBINN}{SBINN}
\newcommand{\PINN}{PINN}
\newcommand{\BIRNN}{BI-RNN}
\newcommand{\FC}{FC}

\newcommand{\MPC}{MPC}
\newcommand{\MIMO}{MIMO}
\newcommand{\loss}[1]{\mathcal{L}^{#1}}

\newcommand{\MSE}{MSE}
\newcommand{\RMSE}{RMSE}
\newcommand{\GOF}{GoF}

\allowdisplaybreaks[1]
\begin{document}

\onecolumn

\title{Integrating Biological-Informed Recurrent Neural Networks for Glucose-Insulin Dynamics Modeling}

\author{
    Stefano De Carli\thanks{S. De Carli, N. Licini, D. Previtali, F. Previdi and A. Ferramosca are with the Department of Management, Information and Production Engineering, University of Bergamo, Via G. Marconi 5, 24044 Dalmine (BG), Italy. {\tt\small stefano.decarli@unibg.it}}
    \And 
    Nicola Licini\footnotemark[1]
    \And 
    Davide Previtali\footnotemark[1]
    \And
    Fabio Previdi\footnotemark[1]
    \And
    Antonio Ferramosca\footnotemark[1]
}

\date{}

\twocolumn[
    \maketitle
    \vspace{-25pt}
    \begin{abstract} 
    	\label{s:abstract}
    	Type 1 Diabetes (T1D) management is a complex task due to many variability factors. Artificial Pancreas (AP) systems have alleviated patient burden by automating insulin delivery through advanced control algorithms. However, the effectiveness of these systems depends on accurate modeling of glucose-insulin dynamics, which traditional mathematical models often fail to capture due to their inability to adapt to patient-specific variations.
    	This study introduces a Biological-Informed Recurrent Neural Network (\BIRNN{}) framework to address these limitations. The \BIRNN{} leverages a Gated Recurrent Units (GRU) architecture augmented with physics-informed loss functions that embed physiological constraints, ensuring a balance between predictive accuracy and consistency with biological principles. The framework is validated using the commercial UVA/Padova simulator, outperforming traditional linear models in glucose prediction accuracy and reconstruction of unmeasured states, even under circadian variations in insulin sensitivity.
    	The results demonstrate the potential of \BIRNN{} for personalized glucose regulation and future adaptive control strategies in AP systems.
    \end{abstract}
    \vspace{\baselineskip}
]
\saythanks


\section{Introduction}
\label{s:introduction}
Managing Type 1 Diabetes (T1D) remains a multifaceted challenge, requiring precise Blood Glucose Level (BGL) regulation through adequate insulin administration. This complexity is heightened by significant variability, both inter-patient and intra-patient, driven by factors such as insulin sensitivity variations, carbohydrate absorption rates, stress, and physical activity~\cite{katsarouTypeDiabetesMellitus2017}.
The advent of the Artificial Pancreas (AP) has revolutionized diabetes management by shifting the responsibility of insulin dosing from patients to automatic control systems. An AP combines a continuous Glucose Monitoring Sensor (CGM), an insulin pump (Continuous Subcutaneous Insulin Infusion, CSII), and a control algorithm to form a hybrid closed-loop system. By analyzing CGM readings, the AP's control algorithm adjusts insulin delivery to maintain BGL within normoglycemic ranges ($70\!\! \leq \!\!\text{BGL} \!\!\leq \!\!140$ [mg/dL]).
Among the various control algorithms developed for AP systems, such as PID controllers~\cite{SteilPID2013}, Fuzzy Logic~\cite{mausethUseFuzzyLogic2013}, and Model Predictive Control (MPC)~\cite{messoriIndividualizedModelPredictive2018}, MPC has emerged as the most promising approach. Its predictive capabilities allow it to anticipate undesired glucose excursions while computing insulin dosages that respect physiological constraints.
MPC relies on a system model to predict future glucose levels and optimize control actions. Typically, these models are based on systems of Ordinary Differential Equations (ODEs) that describe the dynamics of glucose-insulin interactions. Traditional models~\cite{bergman1989, magdelaineLongTermModelGlucose2015, hovorkaNonlinearModelPredictive2004}, provide valuable insights for control applications but often struggle to adapt to unmeasured disturbances, patient-specific variability, or day-to-day fluctuations in physiological parameters.
To overcome these limitations, data-driven approaches have been proposed ~\cite{sonzogniCHoKIBasedMPCBlood2023, SONZOGNI2025100294, polverArtificialPancreasZone2023}, which rely on large datasets to infer system dynamics through input-output relationships. These methods circumvent the complexities of model identification and require minimal prior physiological knowledge. However, they often lack interpretability, computational efficiency, and robustness in clinical applications, particularly when operating with limited or noisy data.
Recent advances in machine learning have opened new possibilities for addressing these challenges. Recurrent Neural Networks (\RNN{}s), specifically architectures such as Long Short-Term Memory (\LSTM{}) networks and Gated Recurrent Units (\GRU{}), excel at capturing temporal dependencies in complex systems. However, purely data-driven methods still lack the reliability and physiological grounding necessary for deployment in real-world scenarios.
In this sense, Physics-Informed Neural Networks (\PINN s) represent a breakthrough by embedding model equations directly into neural network architectures~\cite{RAISSI2019686}. This hybrid approach leverages domain knowledge to constrain the learning process, effectively merging mechanistic modeling with the predictive power of data-driven inference. \PINN s augment neural networks’ ability to generalize while maintaining consistency with known physical principles, making them highly suitable for applications in biomedical engineering.

\paragraphfont{Contributions}
This study proposes a novel Biological-Informed Recurrent Neural Network (\BIRNN{}) framework that integrates machine learning with domain-specific physiological knowledge. Similarly to \PINN s, which embed physics principles to model complex mechanical systems~\cite{RAISSI2019686}, \BIRNN s build upon the System Biological Informed Neural Network (\SBINN{}) framework proposed in~\cite{yazdaniSystemsBiologyInformed2020}, integrating machine learning with domain-specific physiological knowledge by leveraging \RNN{} architectures to model the glucose-insulin dynamics described in~\cite{abuinArtificialPancreasStable2020} as penalties in the loss function. The \BIRNN{} framework achieves high predictive accuracy while preserving adherence to established biological laws. The approach employs a GRU-based architecture augmented with physics-informed losses, balancing data-driven flexibility with mechanistic interpretability. The framework is validated using simulated patient data from the commercial UVA/Padova simulator~\cite{manUVAPADOVAType2014} and compared against the single linear model~\cite{abuinArtificialPancreasStable2020}.

\paragraphfont{Organization}
Here described is this paper workflow. After the introduction in Section \ref{s:introduction}, the methods used to develop the proposed \BIRNN{} will be presented in Section \ref{s:methods}. Section \ref{s:experimental_results} will show the results of the model identification and validation using in-silico simulated scenarios. Finally, Section \ref{s:conclusion} will discuss the implications of our findings and potential future applications in the management of T1D.

\paragraphfont{Notation}
Let $\mathbb{R}$ and $\mathbb{N}$ denote the sets of real and natural numbers, respectively, with $0 \!\in\! \mathbb{N}$. The sets of positive and non-negative real numbers are denoted by $\mathbb{R}_{>0}$ and $\mathbb{R}_{\geq0}$, respectively. For $n, m \!\in\! \mathbb{N}$, $\mathbb{R}^n$ represents the space of $n$-dimensional real column vectors, and $\mathbb{R}^{n \times m}$ is the space of $n \times m$ real matrices. The Hadamard (element-wise) product is denoted by $\circ$. For a set $\mathcal{S}$, its cardinality is written as $\left|\mathcal{S}\right|$. The ceiling function $\lceil \cdot \rceil$ maps a real number to its smallest greater or equal integer. A random subset $\text{rand}(\mathcal{S}, n)$ selects $n$ elements uniformly at random from the set $\mathcal{S}$. Continuous-time signals $s: \mathbb{R}_{\geq0} \to \mathbb{R}$ are expressed as $s(t)$, where $t \!\in\! \mathbb{R}_{\geq0}$ (measured in seconds) is the time variable, and their derivative is denoted by $\dot{s}(t)$. Discrete-time signals $s_k$ are obtained by sampling $s(t)$ as $s_k \!=\! s(kT)$, where $T \!\in\! \mathbb{R}_{>0}$ is the sampling time. $\bar{s} \in \mathbb{R}$ designates the mean of $s(t)$. Boldface denotes vectors. Let $\boldsymbol{0}_n \!\in\! \mathbb{R}^n$ be the $n$-dimensional zero column vector. For $\boldsymbol{x}_k \!=\! [x_{1, k}, \dots, x_{n, k}]^\top \!\in\! \mathbb{R}^n$, define the subvector $\boldsymbol{x}_{r:p,k} \!=\! [x_{r, k}, \dots, x_{p, k}]^\top \!\in\! \mathbb{R}^{p-r+1}$ such that $\boldsymbol{x}_k \!=\! [x_{1, k}, \dots, x_{r-1, k}, \boldsymbol{x}_{r:p,k}^\top, x_{p+1, k}, \dots, x_{n, k}]^\top, 1 \leq r \leq p \leq n$. The component-wise sigmoid and hyperbolic tangent functions of $\boldsymbol{x}_k$ are denoted as $\boldsymbol{y}_k \!=\! \boldsymbol{\sigma}(\boldsymbol{x}_k)$ and $\boldsymbol{z}_k \!=\! \boldsymbol{\tanh}(\boldsymbol{x}_k)$, respectively. These are defined element-wise as $y_{\iota, k} \!=\! \left( 1 + e^{-x_{\iota, k}} \right)^{-1}$ and $z_{\iota, k} \!=\! \left( e^{x_{\iota, k}} - e^{-x_{\iota, k}} \right) \left( e^{x_{\iota, k}} + e^{-x_{\iota, k}} \right)^{-1}$, for all $\iota \!\in\! \{ 1, \dots, n \}$.
\section{Methods}
\label{s:methods}
This Section outlines the key components of the proposed framework, including the mathematical model employed, the architecture and training of the \BIRNN{}, and the loss functions designed to integrate physiological constraints.
\newpage
\subsection{Mathematical model}
\label{ss:mathematical_model}
The mathematical model in this work is based on the proposal in~\cite{ruanModelingDaytoDayVariability2017} and it is represented as:
\begin{subequations}
	\label{eq:mathematical_model}
	\begin{numcases}{}
		\dot{\boldsymbol{y}}(t) = A\boldsymbol{y}(t) + B\boldsymbol{u}(t) + E, \quad \boldsymbol{y}(0) = \boldsymbol{y}_0, \label{eq:state_equation} \\
		\boldsymbol{\gamma}(t) = C\boldsymbol{y}(t), \label{eq:output_equation}
	\end{numcases}
\end{subequations}
with:
\[A=\begin{bmatrix}
	-p_1 & -p_2 & 0 & p_3 & 0\\
	0 & -\sfrac{1}{p_4} & \sfrac{1}{p_4}& 0 & 0\\
	0 & 0 & -\sfrac{1}{p_4} & 0 & 0\\
	0 & 0 & 0 & -\sfrac{1}{p_5} & \sfrac{1}{p_5}\\
	0 & 0 & 0 & 0 & -\sfrac{1}{p_5}\\
\end{bmatrix},\] 
\[
B=\begin{bmatrix}
	0 & 0 \\ 
	0 & 0 \\ 
	\sfrac{1}{p_4} & 0 \\ 
	0 & 0 \\ 
	0 & \sfrac{1}{p_5}
\end{bmatrix} , \
E=\begin{bmatrix}
	p_0 \\ 0 \\ 0 \\ 0 \\ 0
\end{bmatrix} ,
C=\begin{bmatrix}
	1 \\ 0 \\ 0 \\ 0 \\ 0
\end{bmatrix}^\top .
\]
The model consists of 5 states $\boldsymbol{y} = [y_1, \dots, y_5]^\top$ devoted to describing BG concentration $y_1$ [mg/dL] (i.e., the measured output $\gamma$), insulin absorption and action $y_2$ and $y_3$ [U/min], and carbohydrate absorption $y_4$ and $y_5$ [g/min]. The input of the model $\boldsymbol{u}=[u, r]^\top$ is exogenous insulin delivery $u$ [U] and carbohydrate intake $r$ [g]. The dynamics of the states are described using compartmental models, modulated by the set of physiological parameters $p = \left[ p_0, \dots, p_5\right]$ described in Table 1.
These have been previously identified via Regularized Least Squares (RLS) as depicted in~\cite{abuinArtificialPancreasStable2020}.
Moreover, the model consents to estimate the Insulin-On-Board (IOB), i.e. the amount of insulin still to act in the body, as $\text{IOB}(t) = p_4(y_2(t)+y_3(t))$, and the rate of glucose appearance (Ra) in plasma $\text{Ra}(t) = p_3 y_4(t)$ [mg/(dL$\cdot$min)].
Finally, the initial condition, $\boldsymbol{y}(0) = \boldsymbol{y}_0$, is defined at the model's equilibrium state, where $\boldsymbol{y}_0 = [G_b, U_b, U_b, 0, 0]^\top$. Here, $(U_b, G_b)$ represents the values of the basal insulin-glucose pair (i.e. at fasting).

\begin{table}[b]
	\begin{center}
		\caption{Description of the model parameters.}\label{tb:margins}
		\begin{tabular}{@{}ll}
			Parameter & Description \\\hline
			\quad $p_0$ [mg/(dL $\!\!\cdot\!\!$ min)] & Endogenous glucose production (EGP)\\
			\quad $p_1$ [1/min] & Glucose effectiveness\\
			\quad $p_2$ [mg/(dL $\!\!\cdot\!\!$ U)] & Insulin sensitivity\\
			\quad $p_3$ [mg/(dL $\!\!\cdot\!\!$ g)] & Carbohydrate factor\\
			\quad $p_4$ [min] & Insulin absorption time constant\\
			\quad $p_5$ [min] & Meals absorption time constant\\ \hline
		\end{tabular}
	\end{center}
	\label{tab: model param}
\end{table}

\subsection{Biological-Informed RNN (\BIRNN{})}
\label{ss:BIRNN}
A \BIRNN{} extends the System Biological Informed Neural Network (\SBINN{}) framework proposed in~\cite{yazdaniSystemsBiologyInformed2020}. Utilizing \LSTM{} and \GRU{} architectures, \BIRNN{} integrates domain-specific biological constraints while leveraging internal memory mechanisms to capture complex temporal dependencies. These architectures excel in handling Multi-Input Multi-Output (\MIMO{}) systems, making them particularly effective for nonlinear dynamical systems identification, as highlighted in~\cite{ljungDeepLearningSystem2020}. Furthermore, it is shown that the state-space formulation allows for seamless integration into traditional model-based control strategies, such as \MPC{}, as described in~\cite{terziLearningModelPredictive2021}.
Unlike classic \RNN{}s which rely on only the minimization of a data-driven loss, the \BIRNN{} minimizes an augmented loss function with respect to its parameters $\theta$, also accounting for the physiological parameters $p$ in Table 1.
The augmented loss function $\mathcal{L}\left( \theta, p \right)$ employed is:
\begin{equation}
	\label{eq:augmented_loss_function}
	\loss{}\left( \theta, p \right) = \alpha_D \loss{D}\left( \theta \right) + \alpha_B \loss{B}\left( \theta, p \right) + \alpha_A \loss{A}\left( \theta, p \right),
\end{equation}
where:
\begin{itemize}
	\item $\loss{D}\left( \theta\right)$ enforces fit to observed measurements, as in traditional \RNN s.
	\item $\loss{B}\left( \theta, p \right)$ embeds prior knowledge of biological dynamics from the mathematical model employed.
	\item $\loss{A}\left( \theta, p \right)$ encodes auxiliary penalties such as for boundary conditions or additional output enforcements.
	\item $\alpha_j, j \!\in\! \{D, B, A\}$ are the loss function components weights, to treat as new network hyper-parameters.
\end{itemize}

By minimizing $\loss{D}$ and $\loss{B}$, the \BIRNN{} aims to predict the observed outputs while maintaining consistency with the biological underlying system. The inclusion of $\loss{B}$ has a regularization effect, improving the network's generalization under noisy and scarce data.

\subsection{Gated Recurrent Unit (\GRU{}) network model}
\label{ss:gru_network_model}
A \GRU{} network, introduced in~\cite{choLearningPhraseRepresentations2014}, is a simplified version of the \LSTM{} network. It typically consists of $L \!\in\! \mathbb{N}$ layers, each with $n_{\textsuperscript{hu}}^{(l)} \!\in\! \mathbb{N}, l \in \left\{ 1, \dots, L \right\}$, hidden units.
In this work, we focus on single-layer \GRU{} networks and, for simplicity, omit the layer-specific notation $^{(l)}$.
A \GRU{} layer is modeled as a discrete-time nonlinear \MIMO{} dynamical system represented in state-space form with inputs $\boldsymbol{u}_k \!\in\! \mathbb{R}^{n_\textsuperscript{u}}, n_{\text{u}} \!\in\! \mathbb{N}$, and outputs $\boldsymbol{y}_k \!\in\! \mathbb{R}^{n_\textsuperscript{y}}, n_{\text{y}} \!\in\! \mathbb{N}$. The hidden state $\boldsymbol{h}_k \!\in\! \mathbb{R}^{n_{\textsuperscript{hu}}}$ captures the system's internal dynamics. The dynamics of $\boldsymbol{h}_k$ are governed by three gating mechanisms: the reset gate $\boldsymbol{r}_k \!\in\! \mathbb{R}^{n_{\textsuperscript{hu}}}$, the update gate $\boldsymbol{z}_k \!\in\! \mathbb{R}^{n_{\textsuperscript{hu}}}$, and the candidate hidden state $\tilde{\boldsymbol{h}}_k \!\in\! \mathbb{R}^{n_{\textsuperscript{hu}}}$, which collectively control the flow of information. These components are computed as:
\begin{subequations}
	\label{eq:gates_GRU}
	\begin{align}
		\boldsymbol{r}_k &= \boldsymbol{\sigma}\left( W_r {\boldsymbol{u}}_k + R_r \boldsymbol{h}_k + \boldsymbol{b}_r \right),   \\
		\boldsymbol{z}_k &= \boldsymbol{\sigma}\left( W_z {\boldsymbol{u}}_k + R_z \boldsymbol{h}_k + \boldsymbol{b}_z \right),   \\
		\tilde{\boldsymbol{h}}_k &= \boldsymbol{\tanh}\left( W_{\tilde{h}} {\boldsymbol{u}}_k + \boldsymbol{r}_k \circ \left( R_{\tilde{h}} \boldsymbol{h}_k \right) + \boldsymbol{b}_{\tilde{h}} \right),
	\end{align}
\end{subequations}
where $W_j \!\in\! \mathbb{R}^{n_{\textsuperscript{hu}} \!\times\! n_\textsuperscript{u}}$ are the input weights, $R_j \!\in\! \mathbb{R}^{n_{\textsuperscript{hu}} \!\times\! n_{\textsuperscript{hu}}}$ are the recurrent weights, and $\boldsymbol{b}_j \!\in\! \mathbb{R}^{n_{\textsuperscript{hu}}}$ are the biases for $j \!\in\! \{r, z, \tilde{h}\}$. After the \GRU{} network, a Fully Connected (\FC{}) layer is applied to transform the hidden state $\boldsymbol{h}_k$ into the final output. The behavior of the system, governed by the gating mechanisms, is described in the following nonlinear state-space representation:
\begin{subequations}
	\label{eq:GRU_model}
	\begin{numcases}{}
		\label{eq:GRU_hidden_state_update}
		\boldsymbol{h}_{k+1} = \left( 1 - \boldsymbol{z}_k \right) \circ \tilde{\boldsymbol{h}}_k + \boldsymbol{z}_k \circ \boldsymbol{h}_k, \\
		\label{eq:GRU_output_equation}
		\boldsymbol{y}_k = W_y \boldsymbol{h}_{k+1} + \boldsymbol{b}_y, 
	\end{numcases}
\end{subequations}
where \eqref{eq:GRU_output_equation} describes the \FC{} layer mapping the hidden state $\boldsymbol{h}_{k+1}$ to the output vector $\boldsymbol{y}_k$, with $W_y \!\in\! \mathbb{R}^{n_\textsuperscript{y} \!\times\! n_{\textsuperscript{hu}}}$ and $\boldsymbol{b}_y \!\in\! \mathbb{R}^{n_\textsuperscript{y}}$ as the weight matrix and bias vector. In summary, the model in \eqref{eq:GRU_model} relies on a set of parameters
\begin{equation}
	\label{eq:GRU_parameters}
	\theta = \{W_j, R_j, \boldsymbol{b}_j: j \!\in\! \{r, z, \tilde{h}\}\} \cup \{W_y, \boldsymbol{b}_y\},
\end{equation}
which are learned during \GRU{} training. The use of \GRU{}s instead of \LSTM{}s in this application is motivated by their simpler architecture, reducing the number of components to optimize, particularly when paired with augmented loss functions in \BIRNN{}s.

\subsection{Model Training}
\label{ss:model_training}
For model identification, we consider a dataset $\mathcal{D} = \{\mathcal{D}^{(1)}, \dots, \mathcal{D}^{(N\textsubscript{e})}\}$ comprising $N_{\mathrm{e}} \!\in\! \mathbb{N}$ sequences of input-output data. Each sequence $\mathcal{D}^{\text{(e)}} = \{(\boldsymbol{u}_k^{\text{(e)}}, \boldsymbol{y}_k^{\text{(e)}}): k \in \{ 0, \dots, N^{\text{(e)}} - 1\} \}, e \in \left\{ 1, \dots, N\textsubscript{e}\right\}$, is generated by applying input sequences to the system, where $N^{\text{(e)}} \!\in\! \mathbb{N}$ denotes the number of data points in $\mathcal{D}^{\text{(e)}}$. The dataset $\mathcal{D}$ is split into training ($\mathcal{D}_{\mathrm{tr}}$), validation ($\mathcal{D}_{\mathrm{val}}$), and test ($\mathcal{D}_{\mathrm{tst}}$) subsets:
\begin{equation}
	\label{eq:dataset_composition}
	\mathcal{D}_{j} = \{\mathcal{D}^{\text{(e)}}: \text{e} \in \mathcal{I}_{j}\}, \quad j \!\in\! \{\mathrm{tr}, \mathrm{val}, \mathrm{tst}\},
\end{equation}
where $\mathcal{I}_{\mathrm{tr}}, \mathcal{I}_{\mathrm{val}}, \mathcal{I}_{\mathrm{tst}}$ are disjoint subsets of $\{1, \dots, N_{\mathrm{e}}\}$ such that $\mathcal{D}_{\mathrm{tr}} \cup \mathcal{D}_{\mathrm{val}} \cup \mathcal{D}_{\mathrm{tst}} = \mathcal{D}$. The subsets $\mathcal{D}_{\mathrm{tr}}$ and $\mathcal{D}_{\mathrm{val}}$ are used for model estimation, while $\mathcal{D}_{\mathrm{tst}}$ is reserved for performance evaluation. To clarify, the network takes input vectors $\boldsymbol{u}$ and produces output vectors $\boldsymbol{y}$, consistent with the mathematical model in Section \ref{ss:mathematical_model}. For reference, the real $\boldsymbol{y}$ comprises $y_1$ from measured data and $\boldsymbol{y}_\psi$, the four predicted states from the mathematical model in \eqref{eq:mathematical_model}, such that $\boldsymbol{y} = [ y_1, {\boldsymbol{y}_\psi}^\top ]^\top$. Now, we define the training Mean Squared Error (\MSE{}) over the datasets indexed by $\mathcal{I}\textsubscript{tr}$ as:
\begin{equation}
	\label{eq:MSE}
	\mathrm{\MSE{}}\!\left(\theta, \mathcal{I}\textsubscript{tr}\right)\! =\! \frac{1}{\left|\mathcal{I}\textsubscript{tr}\right|} \!\! \sum_{\text{e} \in \mathcal{I}\textsubscript{tr}}\!\!\left[\frac{1}{N^{\text{(e)}}} \!\!\!\!\!\sum_{k = 0}^{N^{\textsubscript{\text{(e)}}} - 1} \!\!\!\!\left( y_{1,k}^{\text{(e)}} \!-\! \hat{y}_{1,k}^{\text{(e)}}\left(\theta\right) \right)^2 \right]\!\!.
\end{equation}
where $\hat{y}_{1,k}^{\text{(e)}}\left(\theta\right)$ is the glucose prediction at time step $k$ computed using the input sequence $\mathbf{u}^{\text{(e)}}$ and the network with parameters $\theta$. The defined \MSE{} will serve as $\loss{D}$ in \eqref{eq:augmented_loss_function}. To incorporate the mathematical model in \eqref{eq:mathematical_model} for $\loss{B}$ computation, the matrices $A$, $B$, and $E$ are discretized using forward Euler discretization as $A_d \!=\! T A \!+\! I$, $B_d \!=\! T B$, and $E_d \!=\! T E$, where $T \!\in\! \mathbb{R}_{>0}$ is the sampling time and $I$ is the identity matrix of appropriate dimension. We then define the sequence biological loss $\ell^B\left(\theta, p \right)^{\text{(e)}}$ as:
\small
\begin{equation}
	\label{eq:individual_biological_loss}
	\ell^B\left(\theta, p \right)^{\text{(e)}} \!\! = \frac{1}{N^{\text{(e)}}} \!\! \!\! \sum_{k = 0}^{N^{\textsubscript{(e)}}-2} \!\! \norm{A_d\hat{\boldsymbol{y}}_{k}^{\text{(e)}}\left(\theta\right)\! + \! B_d\boldsymbol{u}_{k}^{\text{(e)}}\! + \! E_d \! - \! \hat{\boldsymbol{y}}_{k+1}^{\text{(e)}}\left(\theta\right)}{2}^2
\end{equation}
\normalsize
where training $\loss{B}$ follows as:
\begin{equation}
	\label{eq:biological_loss}
	\loss{B}\left(\theta, p, \mathcal{I}\textsubscript{tr}\right) = \frac{1}{\left|\mathcal{I}\textsubscript{tr}\right|} \sum_{\text{e} \in \mathcal{I}\textsubscript{tr}} \ell^B\left(\theta, p \right)^{\text{(e)}}
\end{equation}
Before the calculation of $\loss{A}$, we define the set $\tilde{N}\textsuperscript{(e)}$ as:  
\begin{equation}
	\label{eq:random_set}
	\tilde{N}\textsuperscript{(e)} = \text{rand}(\{1, \dots, N^{\text{(e)}}\}, \lceil \xi N^{\text{(e)}} \rceil),
\end{equation}
where $\xi \!\in\! \left[ 0, 1 \right]$ is the fraction of time steps included. We can now define the sequence auxiliary loss $\ell^A \left(\theta, p, \xi \right)^{\text{(e)}}$ as:
\small
\begin{equation}
	\label{eq:individual_auxiliary_loss}
	\ell^A \left(\theta, p, \xi \right)^{\text{(e)}} = \frac{\ell^A_S\left(\theta, p, \xi \right)^{\text{(e)}} + \ell^A_0\left(\theta, p \right)^{\text{(e)}} + \ell^A_+\left(\theta, \xi \right)^{\text{(e)}}}{3}
\end{equation}
\normalsize
where the computation of the training $\loss{A}$ is:
\begin{equation}
	\label{eq:auxiliary_loss}
	\loss{A}\left(\theta, p, \xi, \mathcal{I}\textsubscript{tr}\right) = \frac{1}{\left|\mathcal{I}\textsubscript{tr}\right|} \sum_{\text{e} \in \mathcal{I}\textsubscript{tr}} \ell^A \left(\theta, p, \xi \right)^{\text{(e)}}
\end{equation}
In \eqref{eq:individual_auxiliary_loss}, the auxiliary loss $\ell^A  \left(\theta, p, \xi \right)^{\text{(e)}}$ comprises three components: the state loss $\ell^A_S\left(\theta, p, \xi \right)^{\text{(e)}}$, which quantifies the discrepancy between the mathematical model states in \eqref{eq:mathematical_model} and the network outputs; the zero loss $\ell^A_0\left(\theta, p \right)^{\text{(e)}}$, capturing the deviation between the model's initial condition and the network's initial outputs; and the positive loss $\ell^A_+\left(\theta, \xi \right)^{\text{(e)}}$, favoring non-negative predicted states. These components are defined as:
\begin{subequations}
	\begin{align}
		\ell&^A_S\left(\theta, p, \xi \right)^{\text{(e)}} = \frac{1}{\tilde{N}\textsuperscript{(e)}} \sum_{k \in \tilde{N}\textsuperscript{(e)}} \norm{\boldsymbol{y}_{\psi,k}^{\text{(e)}}(p) - \hat{\boldsymbol{y}}_{\psi,k}^{\text{(e)}}(\theta)}{2}^2\\
		\ell&^A_0\left(\theta, p \right)^{\text{(e)}} = \norm{\boldsymbol{y}_0(p) - \hat{\boldsymbol{y}}_0^{\text{(e)}}(\theta)}{2}^2\\
		\ell&^A_+\left(\theta, \xi \right)^{\text{(e)}} = \frac{1}{\tilde{N}\textsuperscript{(e)}} \sum_{k \in \tilde{N}\textsuperscript{(e)}} \norm{\max \left( \boldsymbol{0}_4, - \hat{\boldsymbol{y}}_{\psi,k}^{\text{(e)}}(\theta) \right)}{2}^2
		\label{eq:positive_loss}
	\end{align}
\end{subequations}
where $\tilde{N}\textsuperscript{(e)}$ is initialized at every $\loss{A}$ computation. Now that every loss component is defined, the augmented loss $\loss{}$ can be computed as in \eqref{eq:augmented_loss_function}. It is worth noticing that $\ell^A_S\left(\theta, p, \xi \right)^{\text{(e)}}$ and $\ell^A_+\left(\theta, \xi \right)^{\text{(e)}}$ use only a subset of sequence points $\tilde{N}\textsuperscript{(e)}$, which relaxes the promotion of the constraints on $\hat{\boldsymbol{y}}_{\psi}$, reducing rigidity and improving overall convergence of network training.
\\During training, the \BIRNN{} begins with an initial parameter set $\theta_0$, which is iteratively updated using a gradient-based optimizer~\cite{goodfellowDeepLearning2016}. At each step, the parameters $\theta$ are adjusted based on the gradient of the augmented loss function \eqref{eq:augmented_loss_function}, denoted as $\nabla\loss{}(\theta)$, and a learning rate $\eta \!\in\! \mathbb{R}_{\geq0}$. To improve generalization and prevent overfitting, an early stopping strategy is employed. Specifically, every $\kappa_{\mathrm{val}} \!\in\! \mathbb{N}$ iterations, the \MSE{} defined in \eqref{eq:MSE} is computed on the validation set $\mathcal{D}_{\mathrm{val}}$. The current parameter set is stored if the validation error decreases compared to the previous evaluation. Training continues until either a maximum number of iterations $\kappa_{\max} \!\in\! \mathbb{N}$ is reached or the validation error stops improving for a predefined $\rho_{\mathrm{val}} \!\in\! \mathbb{N}$ number of checks. The parameter set achieving the best performance on $\mathcal{D}_{\mathrm{val}}$, denoted as $\theta^*$, is selected as the final model parameters.
\section{Experimental Results}
\label{s:experimental_results}
This Section presents the evaluation of the proposed \BIRNN{}, including dataset generation, training setup, and performance metrics compared to a baseline linear model.

\paragraphfont{Setup}
To generate the datasets including BG values, insulin, and meals, the cohort of 10 adult in-silico patients of the commercial UVA/Padova simulator~\cite{manUVAPADOVAType2014} has been used, specifically the version 1.3.0 by TEG~\cite{t.e.groupDMMS2016}, which implements the circadian variations in insulin sensitivity as in~\cite{toffaninDynamicInsulinBoard2013}. For each patient, datasets were generated using identical inputs to ensure consistency across the cohort.
\\Data for both the training and the testing set ($\mathcal{D}_{\mathrm{tr}}$, $\mathcal{D}_{\mathrm{tst}}$) have been collected following the routine in~\cite{messoriModelIndividualizationArtificial2019} (Table 2, Table 3), while data for the validation set ($\mathcal{D}_{\mathrm{val}}$) followed a 14-day scenario with 3 nominal meals per day (60g at 7:00, 60g at 12:00, and 80g at 18:00), lasting 30 minutes for the first two and 40 for the last. All the events were affected by random variation in meal times ($\pm$ 20 min), size ($\pm$ 20 \%), and duration ($\pm$ 10 min), to ensure the diversity through days of the simulation. Insulin therapy has been defined according to the standard Functional Insulin Treatment (FIT, i.e., continuous basal injection added with postprandial boluses~\cite{howorkaFunctionalInsulinTreatment2012}). Moreover, postprandial boluses were miscalculated by assuming over- or under-estimation of the meal and introducing a delay between 5 and 30 minutes after the meal, to increase the reality of the data collection. The sampling time is $T = 1$ [min].

\paragraphfont{RNN training}
As usual, all data were standardized for network training by subtracting the mean and dividing by the standard deviation. This procedure accounts for the data measured by the simulator and the linear model predictions to be included in the loss as in Section \ref{ss:model_training}. The \BIRNN{} parameters are estimated following the procedure described in Section \ref{ss:model_training}. The learning rate is set to $\eta \!=\! 0.01$, with maximum number of iterations $\kappa_{\max} \!=\! 500$, validation interval $\kappa_{\mathrm{val}} \!=\! 5$, and validation patience $\rho_{\mathrm{val}} \!=\! 20$. The augmented loss function weights in Equation \eqref{eq:augmented_loss_function} are set to $[\alpha_D, \alpha_B, \alpha_A] \!=\! [0.5, 0.25, 0.25]$, prioritizing the nonlinearities learned from measured data while maintaining consistency with the physical model and auxiliary constraints. The sequence fraction $\xi$ in $\loss{A}$ training is set to $0.5$. The \BIRNN{} architecture employs a single-layer \GRU{} network with 96 hidden units $n_{\mathrm{hu}}$, consistent with the hyper-parameters used in~\cite{iaconoPersonalizedLSTMbasedAlarm2023}.
\begin{figure}[!tb]
	\centering
	\includegraphics[width=\columnwidth]{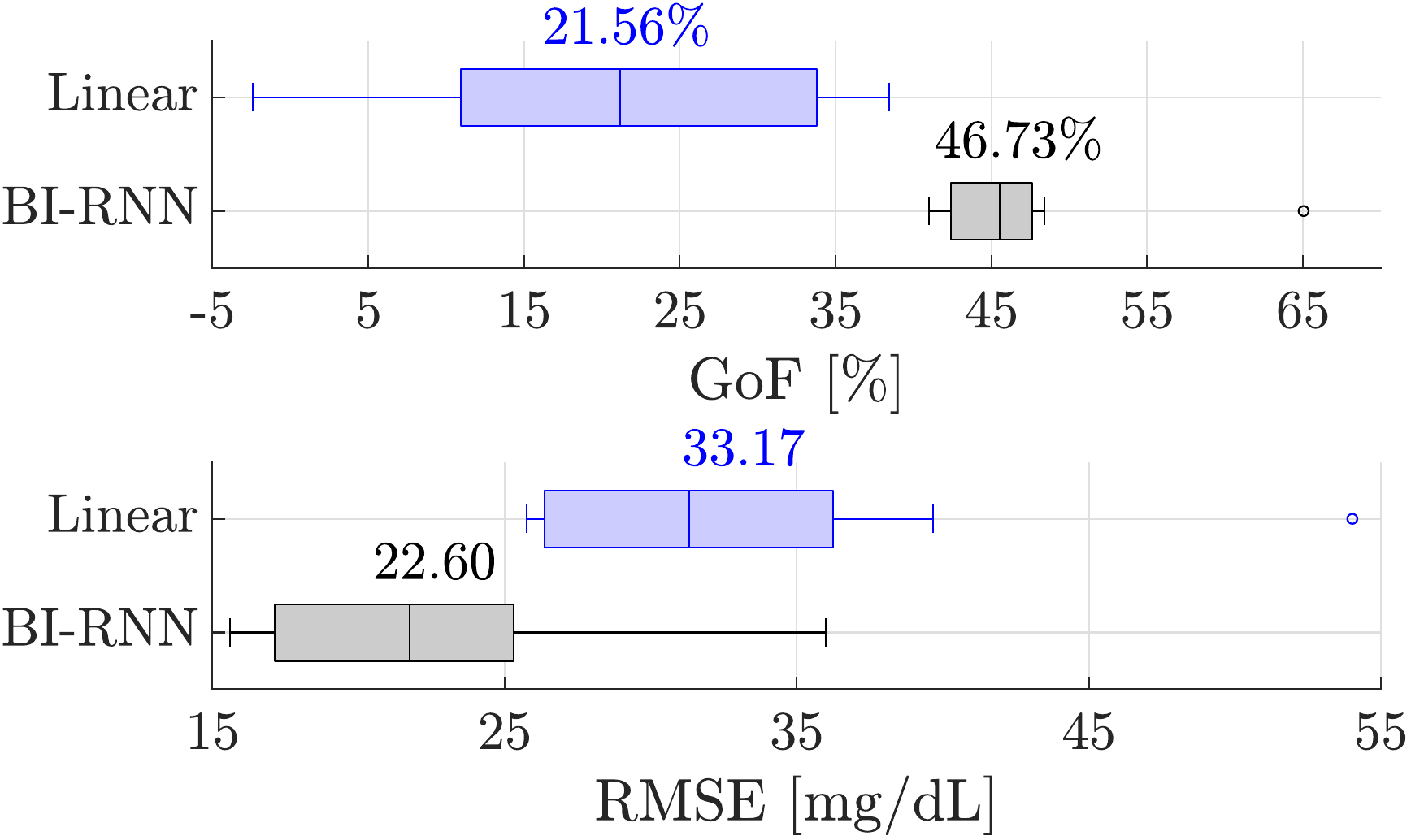}
	\caption{
		\label{fig:fit_rmse}
		Performance metrics comparing the \BIRNN{} and the linear model proposed in~\cite{abuinArtificialPancreasStable2020} in simulating the test scenario. The evaluation is performed on the 10 adult patients of the commercial version of the UVA/Padova simulator. The selected network performs particularly well for adult patient number 9, with a GoF of 65.04\%.  
	}
\end{figure}
\begin{figure*}[!tb]
	\centering
	\includegraphics[width=0.49\textwidth]{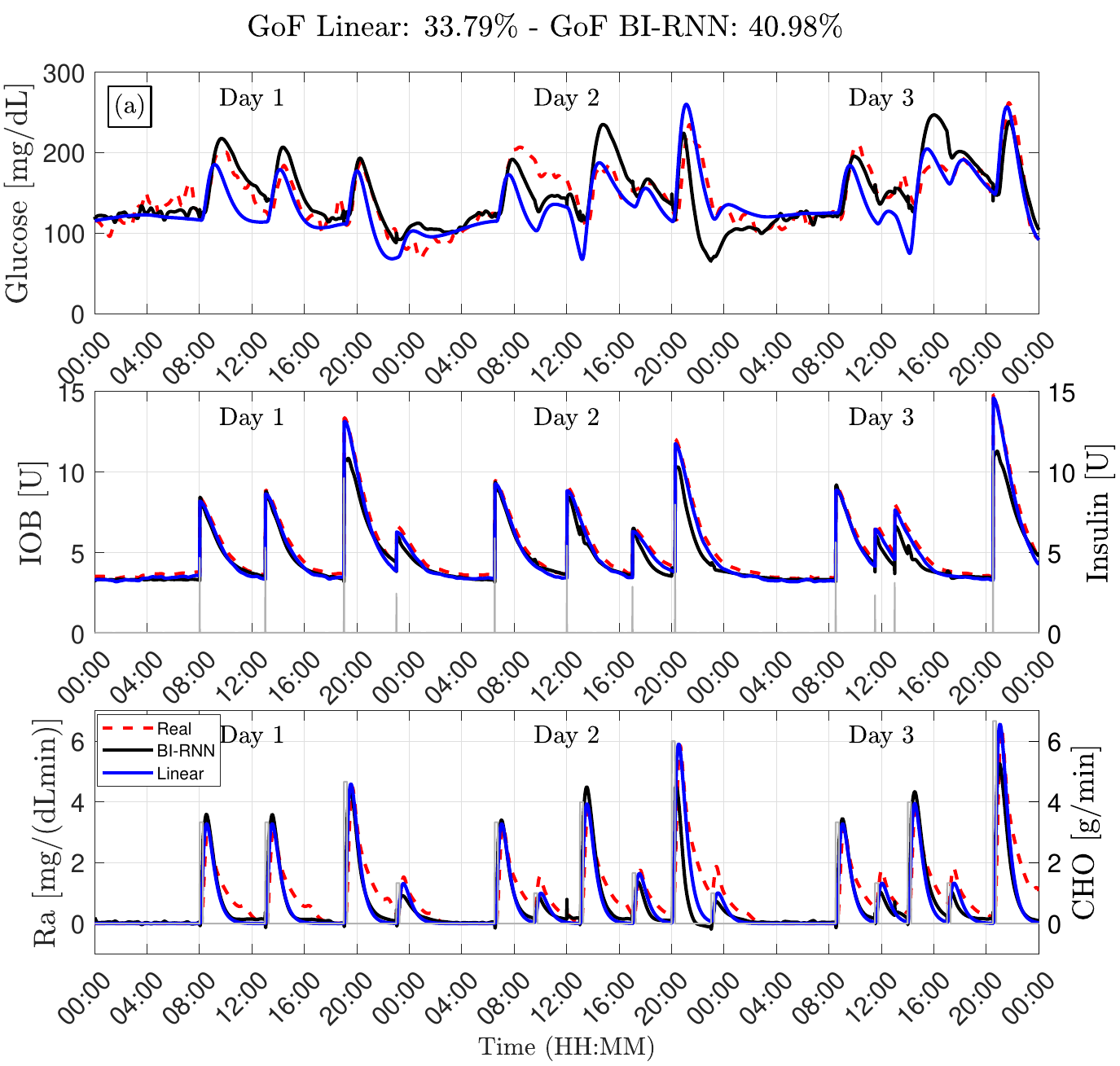}
	\includegraphics[width=0.49\textwidth]{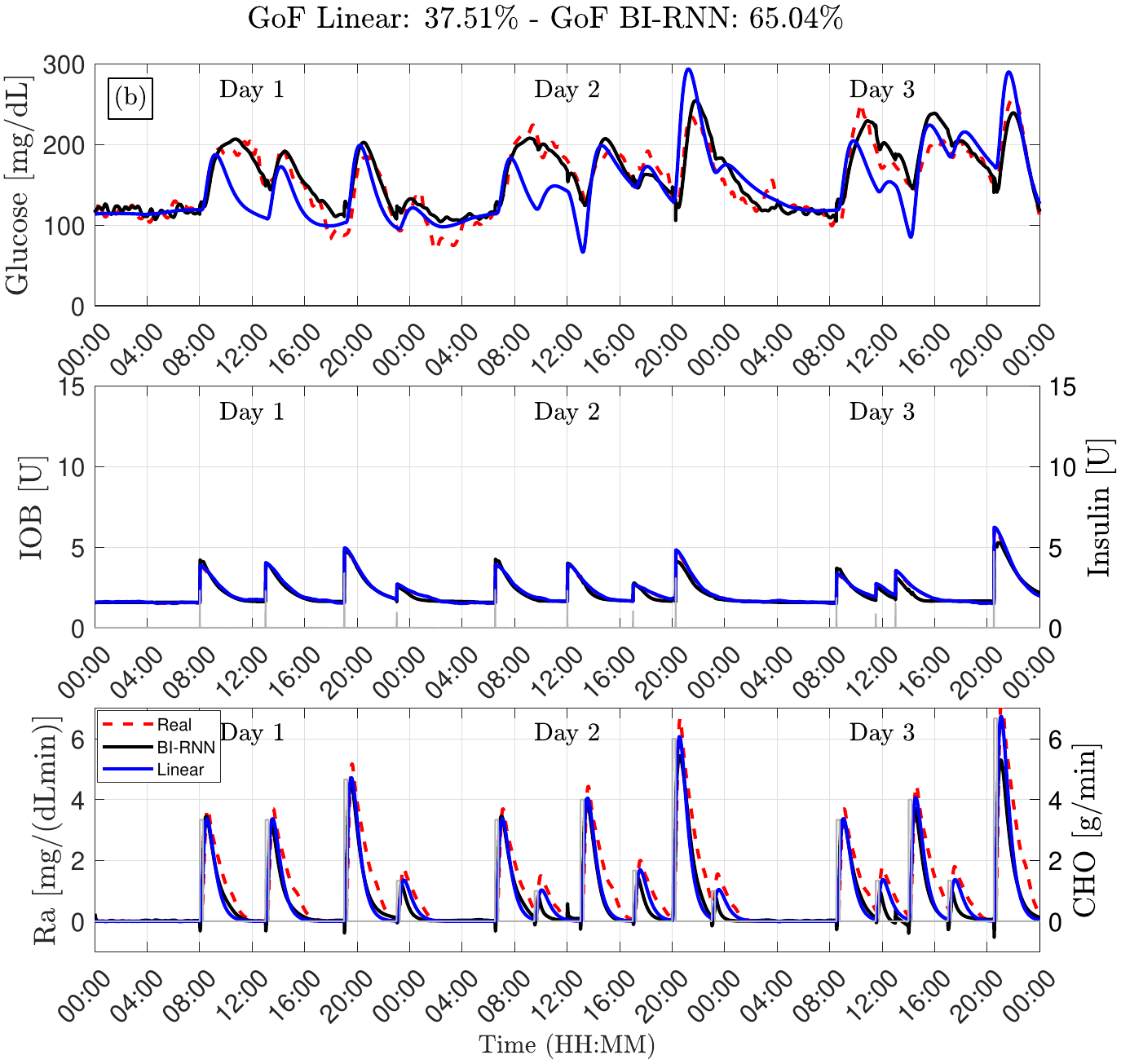}
	\caption{
		\label{fig:predictions}
		Testing scenario comparison of BGL, IOB, and Ra for the adult patient number 6 (a. - worst \BIRNN{} result) and adult patient number 9 (b. - best \BIRNN{} result) of the UVA/Padova in-silico commercial cohort. The red dotted line represents the ground truth, based on the nonlinear time-variant model from UVA/Padova. The black line corresponds to predictions obtained using the proposed \BIRNN{}, while the blue line reflects predictions from the linear model described in~\cite{abuinArtificialPancreasStable2020}. Notably, the \BIRNN{} achieves a closer fit to real glucose data while maintaining accurate estimations for both IOB and Ra.
	}
\end{figure*}

\paragraphfont{Metrics for Evaluation}
The employed metrics to evaluate the model performance on a general test sequence $\mathcal{D}^{\text{(e)}} \!\in\! \mathcal{D}_{\mathrm{tst}}$ are respectively the Root Mean Squared Error (\RMSE{}) and the Goodness of Fit (\GOF{}), both in terms of glucose prediction. While the $\text{\RMSE{}}^{\text{(e)}}$ is simply computed as the square root of $\text{\MSE{}}^{\text{(e)}}$ with only the designated test sequence $\mathcal{D}^{\text{(e)}}$, the $\GOF{}^{\text{(e)}}$ is computed as:
\begin{equation}
	\label{es:goodness_of_fit}
	\mathrm{\GOF{}}^{\text{(e)}}\left(\theta\right) = \frac{100}{N^{\text{(e)}}} \sum_{k = 0}^{N^{\textsubscript{(e)}} - 1} \left( 1 - \frac{\left| y_{1,k}^{\text{(e)}} - \hat{y}_{1,k}^{\text{(e)}}\left(\theta\right) \right|}{\left| y_{1,k}^{\text{(e)}} - \bar{y}_1^{\text{(e)}}\left(\theta\right) \right|} \right).
\end{equation}

\paragraphfont{Results}
The proposed \BIRNN{} demonstrates good performance in simulating key physiological variables for the T1DM scenario, outperforming the linear model proposed in~\cite{abuinArtificialPancreasStable2020} in both metrics employed, as visible in Figure \ref{fig:fit_rmse}. The overall GoF, evaluated by means of \eqref{es:goodness_of_fit} over the entire 10 patients cohort, has a median value of 46.73, distributed between 42.38 (25th percentile) and 47.62 (75th percentile). Notably, for adult patient number 9, the \BIRNN{} achieves a GoF of 65.04\%, reflecting its ability to align closely with the simulator's nonlinear, time-varying ground truth model, as visible in Figure \ref{fig:predictions}. 
Slight deviations occur in the state predictions, also comprehending occasional values below zero given by the network nonlinearities. This inconsistent behavior in the predicted states with the imposed loss in Equation \eqref{eq:positive_loss} arises because there is a trade-off between the different components of the augmented loss $\loss{}$ given by the weights $\alpha_j, j \in \{D, B, A\}$, which mostly favors the fit of the observed measurement. 
However, the overall trajectory of the predictions, including BGL, IOB, and Ra (computed as in Section \ref{ss:mathematical_model}), aligns well with the physiological patterns. This is noteworthy considering the network's good ability to make accurate inferences of model states, making the predictions not only reliable but also feasible for control applications~\cite{gondhalekarVelocityweightingVelocitypenaltyMPC2018, abuinPulsatileZoneMPC2024}.
\\As a side note, a single-layer \LSTM{} network with identical tuning and training strategy as the \GRU{} network was also tested for each patient. However, it consistently produced suboptimal results, at best matching the performance of the linear model.
\section{Conclusion}
\label{s:conclusion}
This work introduces a novel \BIRNN{} framework tailored for modeling and predicting glucose-insulin dynamics in T1D management. By promoting physiological constraints directly in a \GRU-based architecture, the framework achieves a unique balance between predictive accuracy and adherence to established biological dynamics. Validation using simulated patient data highlighted its significant advantages over traditional linear models, particularly in glucose prediction and the reconstruction of unmeasured states.
The \BIRNN{} framework’s capability to capture underlying physiological dynamics makes it particularly promising for personalized control strategies, such as those employed in AP systems. This approach allows for better alignment with individual patient profiles, offering a pathway toward more effective and patient-specific glucose regulation. The results also suggest that further tuning and personalization of the \BIRNN{} configuration, possibly through transfer learning or additional training on individual patient datasets, could enhance its performance and robustness.
\\Several future directions for this work emerge. First, extending the framework to incorporate additional physiological factors, such as the impact of physical activity~\cite{schiavonSilicoOptimizationBasal2013, liciniArtificialPancreasStable2024}, would broaden its applicability in real-world scenarios. In addition, adhering to best practices in machine learning and diabetes research - such as incorporating a more diverse cohort by including pediatric patients and adolescents, using standardized data sets, and benchmarking against published algorithms~\cite{jacobsArtificialIntelligenceMachine2024} - is critical for ensuring the reliability and effectiveness of future advances.
Finally, integrating \BIRNN{} into existing control frameworks, such as MPC strategies~\cite{castilloDeepNeuralNetwork2023}, could pave the way for the next generation of closed-loop AP systems, possibly improving glycemic control and potentially enhancing the quality of life for people living with T1D.

\section*{Acknowledgement}
This work was funded by the National Plan for NRRP Complementary Investments (PNC, established with the decree-law 6 May 2021, n. 59, converted by law n. 101 of 2021) in the call for the funding of research initiatives for technologies and innovative trajectories in the health and care sectors (Directorial Decree n. 931 of 06-06-2022) - project n. PNC0000003 - AdvaNced Technologies for Human-centrEd Medicine (project acronym: ANTHEM).

\bibliographystyle{plain}
{
\bibliography{biblio} 

@article{katsarouTypeDiabetesMellitus2017,
  title = {Type 1 Diabetes Mellitus},
  author = {Katsarou, Anastasia and Gudbj{\"o}rnsdottir, Soffia and Rawshani, Araz and Dabelea, Dana and Bonifacio, Ezio and Anderson, Barbara J. and Jacobsen, Laura M. and Schatz, Desmond A. and Lernmark, Ake},
  year = {2017},
  month = mar,
  journal = {Nature Reviews Disease Primers},
  volume = {3},
  number = {1},
  pages = {17016},
  issn = {2056-676X},
  doi = {10.1038/nrdp.2017.16}
}

@article{SteilPID2013,
  title = {Algorithms for a Closed-Loop Artificial Pancreas: {{The}} Case for Proportional-Integral-Derivative Control},
  author = {Steil, Garry M.},
  year = {2013},
  journal = {Journal of Diabetes Science and Technology},
  volume = {7},
  number = {6},
  eprint = {https://doi.org/10.1177/193229681300700623},
  pages = {1621--1631},
  doi = {10.1177/193229681300700623}
}

@article{mausethUseFuzzyLogic2013,
  title = {Use of a ``{{Fuzzy Logic}}'' {{Controller}} in a {{Closed-Loop Artificial Pancreas}}},
  author = {Mauseth, Richard and Hirsch, Irl B. and Bollyky, Jennifer and Kircher, Robert and Matheson, Don and Sanda, Srinath and Greenbaum, Carla},
  year = {2013},
  month = aug,
  journal = {Diabetes Technology \& Therapeutics},
  volume = {15},
  number = {8},
  pages = {628--633},
  publisher = {Mary Ann Liebert, Inc., publishers},
  issn = {1520-9156},
  doi = {10.1089/dia.2013.0036},
  urldate = {2024-12-05}
}

@article{messoriIndividualizedModelPredictive2018,
  title = {Individualized Model Predictive Control for the Artificial Pancreas: {{In}} Silico Evaluation of Closed-Loop Glucose Control},
  shorttitle = {Individualized Model Predictive Control for the Artificial Pancreas},
  author = {Messori, Mirko and Paolo Incremona, Gian and Cobelli, Claudio and Magni, Lalo},
  year = {2018},
  month = feb,
  journal = {IEEE Control Systems},
  volume = {38},
  number = {1},
  pages = {86--104},
  issn = {1066-033X, 1941-000X},
  doi = {10.1109/MCS.2017.2766314},
  urldate = {2024-12-05},
  copyright = {https://ieeexplore.ieee.org/Xplorehelp/downloads/license-information/IEEE.html},
  langid = {english}
}

@article{bergman1989,
    author = {Bergman, Richard N},
    title = {Toward Physiological Understanding of Glucose Tolerance: Minimal-Model Approach},
    journal = {Diabetes},
    volume = {38},
    number = {12},
    pages = {1512-1527},
    year = {1989},
    month = {12},
    issn = {0012-1797},
    doi = {10.2337/diab.38.12.1512},
    url = {https://doi.org/10.2337/diab.38.12.1512}
}

@article{magdelaineLongTermModelGlucose2015,
  title = {A {{Long-Term Model}} of the {{Glucose}}--{{Insulin Dynamics}} of {{Type}} 1 {{Diabetes}}},
  author = {Magdelaine, Nicolas and Chaillous, Lucy and Guilhem, Isabelle and Poirier, Jean-Yves and Krempf, Michel and Moog, Claude H. and Le Carpentier, Eric},
  year = {2015},
  month = jun,
  journal = {IEEE Transactions on Biomedical Engineering},
  volume = {62},
  number = {6},
  pages = {1546--1552},
  issn = {0018-9294, 1558-2531},
  doi = {10.1109/TBME.2015.2394239},
  urldate = {2024-05-06},
  copyright = {https://ieeexplore.ieee.org/Xplorehelp/downloads/license-information/EU.html},
  langid = {english}
}

@article{hovorkaNonlinearModelPredictive2004,
  title = {Nonlinear Model Predictive Control of Glucose Concentration in Subjects with Type 1 Diabetes},
  author = {Hovorka, Roman and Canonico, Valentina and Chassin, Ludovic J and Haueter, Ulrich and {Massi-Benedetti}, Massimo and Federici, Marco Orsini and Pieber, Thomas R and Schaller, Helga C and Schaupp, Lukas and Vering, Thomas and Wilinska, Malgorzata E},
  year = {2004},
  month = aug,
  journal = {Physiological Measurement},
  volume = {25},
  number = {4},
  pages = {905--920},
  issn = {0967-3334, 1361-6579},
  doi = {10.1088/0967-3334/25/4/010},
  urldate = {2024-05-06},
  langid = {english}
}

@article{sonzogniCHoKIbasedMPCBlood2023,
  title = {{{CHoKI-based MPC}} for Blood Glucose Regulation in Artificial {{Pancreas}}},
  author = {Sonzogni, Beatrice and Manzano, Jos{\'e} Mar{\'i}a and Polver, Marco and Previdi, Fabio and Ferramosca, Antonio},
  year = {2023},
  journal = {IFAC-PapersOnLine},
  volume = {56},
  number = {2},
  pages = {9672--9677},
  issn = {24058963},
  doi = {10.1016/j.ifacol.2023.10.276},
  urldate = {2024-09-20},
  langid = {english}
}

@article{SONZOGNI2025100294,
title = {CHoKI-based MPC for blood glucose regulation in Artificial Pancreas},
journal = {IFAC Journal of Systems and Control},
volume = {31},
pages = {100294},
year = {2025},
issn = {2468-6018},
doi = {https://doi.org/10.1016/j.ifacsc.2024.100294},
url = {https://www.sciencedirect.com/science/article/pii/S2468601824000555},
author = {Beatrice Sonzogni and José María Manzano and Marco Polver and Fabio Previdi and Antonio Ferramosca}
}

@article{polverArtificialPancreasZone2023,
  title = {Artificial {{Pancreas}} under a {{Zone Model Predictive Control}} Based on {{Gaussian Process}} Models: Toward the Personalization of the Closed Loop},
  shorttitle = {Artificial {{Pancreas}} under a {{Zone Model Predictive Control}} Based on {{Gaussian Process}} Models},
  author = {Polver, Marco and Sonzogni, Beatrice and Mazzoleni, Mirko and Previdi, Fabio and Ferramosca, Antonio},
  year = {2023},
  journal = {IFAC-PapersOnLine},
  volume = {56},
  number = {2},
  pages = {9642--9647},
  issn = {24058963},
  doi = {10.1016/j.ifacol.2023.10.271},
  urldate = {2024-09-18},
  langid = {english}
}

@article{RAISSI2019686,
title = {Physics-informed neural networks: A deep learning framework for solving forward and inverse problems involving nonlinear partial differential equations},
journal = {Journal of Computational Physics},
volume = {378},
pages = {686-707},
year = {2019},
issn = {0021-9991},
doi = {https://doi.org/10.1016/j.jcp.2018.10.045},
url = {https://www.sciencedirect.com/science/article/pii/S0021999118307125},
author = {M. Raissi and P. Perdikaris and G.E. Karniadakis}
}

@article{yazdaniSystemsBiologyInformed2020,
  title = {Systems Biology Informed Deep Learning for Inferring Parameters and Hidden Dynamics},
  author = {Yazdani, Alireza and Lu, Lu and Raissi, Maziar and Karniadakis, George Em},
  editor = {Hatzimanikatis, Vassily},
  year = {2020},
  month = nov,
  journal = {PLOS Computational Biology},
  volume = {16},
  number = {11},
  pages = {e1007575},
  issn = {1553-7358},
  doi = {10.1371/journal.pcbi.1007575},
  urldate = {2024-11-25},
  langid = {english}
}

@article{abuinArtificialPancreasStable2020,
  title = {Artificial Pancreas under Stable Pulsatile {{MPC}}: {{Improving}} the Closed-Loop Performance},
  shorttitle = {Artificial Pancreas under Stable Pulsatile {{MPC}}},
  author = {Abuin, P. and Rivadeneira, P.S. and Ferramosca, A. and Gonz{\'a}lez, A.H.},
  year = {2020},
  month = aug,
  journal = {Journal of Process Control},
  volume = {92},
  pages = {246--260},
  issn = {09591524},
  doi = {10.1016/j.jprocont.2020.06.009},
  urldate = {2024-05-06},
  langid = {english}
}

@article{manUVAPADOVAType2014,
  title = {The {{UVA}}/{{PADOVA Type}} 1 {{Diabetes Simulator}}: {{New Features}}},
  shorttitle = {The {{UVA}}/{{PADOVA Type}} 1 {{Diabetes Simulator}}},
  author = {Man, Chiara Dalla and Micheletto, Francesco and Lv, Dayu and Breton, Marc and Kovatchev, Boris and Cobelli, Claudio},
  year = {2014},
  month = jan,
  journal = {Journal of Diabetes Science and Technology},
  volume = {8},
  number = {1},
  pages = {26--34},
  issn = {1932-2968, 1932-2968},
  doi = {10.1177/1932296813514502},
  urldate = {2024-06-18},
  langid = {english}
}

@article{ruanModelingDaytoDayVariability2017,
  title = {Modeling {{Day-to-Day Variability}} of {{Glucose}}--{{Insulin Regulation Over}} 12-{{Week Home Use}} of {{Closed-Loop Insulin Delivery}}},
  author = {Ruan, Yue and Wilinska, Malgorzata E. and Thabit, Hood and Hovorka, Roman},
  year = {2017},
  month = jun,
  journal = {IEEE Transactions on Biomedical Engineering},
  volume = {64},
  number = {6},
  pages = {1412--1419},
  issn = {0018-9294, 1558-2531},
  doi = {10.1109/TBME.2016.2590498},
  urldate = {2024-05-06},
  copyright = {https://creativecommons.org/licenses/by/3.0/legalcode},
  langid = {english}
}

@book{goodfellowDeepLearning2016,
  title = {Deep Learning},
  author = {Goodfellow, Ian and Bengio, Yoshua and Courville, Aaron},
  year = {2016},
  series = {Adaptive Computation and Machine Learning},
  publisher = {The MIT press},
  address = {Cambridge, Mass},
  isbn = {978-0-262-03561-3},
  langid = {english},
  lccn = {006.31}
}

@misc{t.e.groupDMMS2016,
  title = {{{DMMS}}.{{R}}},
  author = {T. E. Group},
  year = {2016},
  howpublished = {T. E. Group}
}

@article{toffaninDynamicInsulinBoard2013,
  title = {Dynamic {{Insulin}} on {{Board}}: {{Incorporation}} of {{Circadian Insulin Sensitivity Variation}}},
  shorttitle = {Dynamic {{Insulin}} on {{Board}}},
  author = {Toffanin, Chiara and Zisser, Howard and Doyle, Francis J. and Dassau, Eyal},
  year = {2013},
  month = jul,
  journal = {Journal of Diabetes Science and Technology},
  volume = {7},
  number = {4},
  pages = {928--940},
  issn = {1932-2968, 1932-2968},
  doi = {10.1177/193229681300700415},
  urldate = {2024-05-06},
  langid = {english}
}

@article{ljungDeepLearningSystem2020,
  title = {Deep {{Learning}} and {{System Identification}}},
  author = {Ljung, Lennart and Andersson, Carl and Tiels, Koen and Sch{\"o}n, Thomas B.},
  year = {2020},
  journal = {IFAC-PapersOnLine},
  volume = {53},
  number = {2},
  pages = {1175--1181},
  issn = {24058963},
  doi = {10.1016/j.ifacol.2020.12.1329},
  urldate = {2024-10-01},
  langid = {english}
}

@article{terziLearningModelPredictive2021,
  title = {Learning Model Predictive Control with Long Short-term Memory Networks},
  author = {Terzi, Enrico and Bonassi, Fabio and Farina, Marcello and Scattolini, Riccardo},
  year = {2021},
  month = dec,
  journal = {International Journal of Robust and Nonlinear Control},
  volume = {31},
  number = {18},
  pages = {8877--8896},
  issn = {1049-8923, 1099-1239},
  doi = {10.1002/rnc.5519},
  urldate = {2024-11-25},
  langid = {english}
}

@inproceedings{choLearningPhraseRepresentations2014,
  title = {Learning {{Phrase Representations}} Using {{RNN Encoder}}--{{Decoder}} for {{Statistical Machine Translation}}},
  booktitle = {Proceedings of the 2014 {{Conference}} on {{Empirical Methods}} in {{Natural Language Processing}} ({{EMNLP}})},
  author = {Cho, Kyunghyun and Van Merrienboer, Bart and Gulcehre, Caglar and Bahdanau, Dzmitry and Bougares, Fethi and Schwenk, Holger and Bengio, Yoshua},
  year = {2014},
  pages = {1724--1734},
  publisher = {Association for Computational Linguistics},
  address = {Doha, Qatar},
  doi = {10.3115/v1/D14-1179},
  urldate = {2024-09-24},
  langid = {english}
}

@article{messoriModelIndividualizationArtificial2019,
  title = {Model Individualization for Artificial Pancreas},
  author = {Messori, Mirko and Toffanin, Chiara and Del Favero, Simone and De Nicolao, Giuseppe and Cobelli, Claudio and Magni, Lalo},
  year = {2019},
  month = apr,
  journal = {Computer Methods and Programs in Biomedicine},
  volume = {171},
  pages = {133--140},
  issn = {01692607},
  doi = {10.1016/j.cmpb.2016.06.006},
  urldate = {2024-05-06},
  langid = {english}
}

@book{howorkaFunctionalInsulinTreatment2012,
  title = {Functional {{Insulin Treatment}}: {{Principles}}, {{Teaching Approach}} and {{Practice}}},
  author = {Howorka, Kinga},
  year = {2012},
  publisher = {Springer Science \& Business Media}
}

@article{iaconoPersonalizedLSTMbasedAlarm2023,
  title = {Personalized {{LSTM-based}} Alarm Systems for Hypoglycemia and Hyperglycemia Prevention},
  author = {Iacono, Francesca and Magni, Lalo and Toffanin, Chiara},
  year = {2023},
  month = sep,
  journal = {Biomedical Signal Processing and Control},
  volume = {86},
  pages = {105167},
  issn = {17468094},
  doi = {10.1016/j.bspc.2023.105167},
  urldate = {2024-11-25},
  langid = {english}
}

@article{gondhalekarVelocityweightingVelocitypenaltyMPC2018,
  title = {Velocity-Weighting \& Velocity-Penalty {{MPC}} of an Artificial Pancreas: {{Improved}} Safety \& Performance},
  shorttitle = {Velocity-Weighting \& Velocity-Penalty {{MPC}} of an Artificial Pancreas},
  author = {Gondhalekar, Ravi and Dassau, Eyal and Doyle, Francis J.},
  year = {2018},
  month = may,
  journal = {Automatica},
  volume = {91},
  pages = {105--117},
  issn = {00051098},
  doi = {10.1016/j.automatica.2018.01.025},
  urldate = {2024-08-02},
  langid = {english}
}

@article{abuinPulsatileZoneMPC2024,
  title = {Pulsatile {{Zone MPC}} with Asymmetric Stationary Cost for Artificial Pancreas Based on a Non-Standard {{IOB}} Constraint},
  author = {Abuin, Pablo and Ferramosca, Antonio and Toffanin, Chiara and Magni, Lalo and Gonz{\'a}lez, Alejandro H.},
  year = {2024},
  month = apr,
  journal = {Journal of Process Control},
  volume = {136},
  pages = {103191},
  issn = {09591524},
  doi = {10.1016/j.jprocont.2024.103191},
  urldate = {2024-05-06},
  langid = {english}
}

@article{schiavonSilicoOptimizationBasal2013,
  title = {{\emph{In }}{{{\emph{Silico}}}} {{Optimization}} of {{Basal Insulin Infusion Rate}} during {{Exercise}}: {{Implication}} for {{Artificial Pancreas}}},
  shorttitle = {{\emph{In }}{{{\emph{Silico}}}} {{Optimization}} of {{Basal Insulin Infusion Rate}} during {{Exercise}}},
  author = {Schiavon, Michele and Man, Chiara Dalla and Kudva, Yogish C. and Basu, Ananda and Cobelli, Claudio},
  year = {2013},
  month = nov,
  journal = {Journal of Diabetes Science and Technology},
  volume = {7},
  number = {6},
  pages = {1461--1469},
  issn = {1932-2968, 1932-2968},
  doi = {10.1177/193229681300700606},
  urldate = {2024-05-06},
  langid = {english}
}

@inproceedings{liciniArtificialPancreasStable2024,
  title = {Artificial {{Pancreas}} under Stable Pulsatile {{Model Predictive Control}}: Including the {{Physical Activity}} Effect},
  booktitle = {Papers Accepted for Publication in the 63rd {{Conference}} on {{Decision}} and {{Control}} ({{CDC}}), 2024.},
  author = {Licini, Nicola and Sonzogni, Beatrice and Abuin, Pablo and Previdi, Fabio and Gonzalez, Alejandro H and Ferramosca, Antonio},
  year = {2024},
  month = dec,
  publisher = {IEEE},
  address = {Milan, Italy},
  langid = {english}
}

@article{jacobsArtificialIntelligenceMachine2024,
  title = {Artificial {{Intelligence}} and {{Machine Learning}} for {{Improving Glycemic Control}} in {{Diabetes}}: {{Best Practices}}, {{Pitfalls}}, and {{Opportunities}}},
  shorttitle = {Artificial {{Intelligence}} and {{Machine Learning}} for {{Improving Glycemic Control}} in {{Diabetes}}},
  author = {Jacobs, Peter G. and Herrero, Pau and Facchinetti, Andrea and Vehi, Josep and Kovatchev, Boris and Breton, Marc D. and Cinar, Ali and Nikita, Konstantina S. and Doyle, Francis J. and Bondia, Jorge and Battelino, Tadej and Castle, Jessica R. and Zarkogianni, Konstantia and Narayan, Rahul and {Mosquera-Lopez}, Clara},
  year = {2024},
  journal = {IEEE Reviews in Biomedical Engineering},
  volume = {17},
  pages = {19--41},
  issn = {1937-3333, 1941-1189},
  doi = {10.1109/RBME.2023.3331297},
  urldate = {2024-05-06},
  copyright = {https://creativecommons.org/licenses/by/4.0/legalcode},
  langid = {english}
}

@article{castilloDeepNeuralNetwork2023,
  title = {Deep {{Neural Network Architectures}} for an {{Embedded MPC Implementation}}: {{Application}} to an {{Automated Insulin Delivery System}}},
  shorttitle = {Deep {{Neural Network Architectures}} for an {{Embedded MPC Implementation}}},
  author = {Castillo, A. and {Villa-Tamayo}, M.F. and Pryor, E. and {Garcia-Tirado}, J. and Colmegna, P. and Breton, M.},
  year = {2023},
  journal = {IFAC-PapersOnLine},
  volume = {56},
  number = {2},
  pages = {11521--11526},
  issn = {24058963},
  doi = {10.1016/j.ifacol.2023.10.443},
  urldate = {2025-01-09},
  langid = {english}
}
}

\end{document}